\pdfoutput=1
\documentclass[11pt,a4paper,table]{article}
\usepackage[hyperref]{emnlp2018}
\usepackage{natbib}
\usepackage{times}
\usepackage{latexsym}
\usepackage{amsmath}
\usepackage{amssymb}
\usepackage{amsthm}
\usepackage{booktabs}
\usepackage{graphicx}
\usepackage{array}

\usepackage{caption}
\usepackage{setspace}
\usepackage{subcaption}
\usepackage{multirow}
\usepackage{algorithm}
\usepackage{algorithmic}
\usepackage{ifthen}
\usepackage{color}
\usepackage[utf8]{inputenc}
\usepackage{url}
\usepackage{multicol}
\usepackage{avm}
\usepackage{tikz}
\usepackage{booktabs}
\usepackage{xcolor}
\usepackage{csquotes}
\usepackage{soul}
\usepackage{float}
\usetikzlibrary{fit}
\usetikzlibrary{patterns}

\usepackage{enumitem}
\usepackage{tabularx}

\aclfinalcopy 

\title{Textual Analogy Parsing: What's Shared and \\ What's Compared among Analogous Facts}
\author{%
Matthew Lamm$^{1,3}$ \quad Arun Tejasvi Chaganty$^{2,3}$\Thanks{Author contributed significantly.} \\ \textbf{Christopher D. Manning}$^{1,2,3}$ \quad \textbf{Dan Jurafsky}$^{1,2,3}$ \quad 
\textbf{Percy Liang}$^{2,3}$ \\
	$^1$Stanford Linguistics \quad
   $^2$Stanford Computer Science \quad
   $^3$Stanford NLP Group \\
   {\tt \{mlamm,jurafsky\}@stanford.edu } \\
   {\tt  \{chaganty,manning,pliang\}@cs.stanford.edu} }
\date{}

\providecommand\bx{\ensuremath{\mathbf{x}}}

\newcommand\eqdef{\ensuremath{\stackrel{\rm def}{=}}} 

\newcommand\refsec[1]{Section~\ref{sec:#1}}

\newcommand\reffig[1]{Figure~\ref{fig:#1}}

\newcommand\reftab[1]{Table~\ref{tab:#1}}

\ifthenelse{\isundefined{\definition}}{\newtheorem{definition}{Definition}}{}
\ifthenelse{\isundefined{\assumption}}{\newtheorem{assumption}{Assumption}}{}
\ifthenelse{\isundefined{\hypothesis}}{\newtheorem{hypothesis}{Hypothesis}}{}
\ifthenelse{\isundefined{\proposition}}{\newtheorem{proposition}{Proposition}}{}
\ifthenelse{\isundefined{\theorem}}{\newtheorem{theorem}{Theorem}}{}
\ifthenelse{\isundefined{\lemma}}{\newtheorem{lemma}{Lemma}}{}
\ifthenelse{\isundefined{\corollary}}{\newtheorem{corollary}{Corollary}}{}
\ifthenelse{\isundefined{\alg}}{\newtheorem{alg}{Algorithm}}{}
\ifthenelse{\isundefined{\example}}{\newtheorem{example}{Example}}{}
\avmfont{\sc\tiny}
\avmvalfont{\rm\tiny}

\definecolor{CMpurple}{rgb}{0.6,0.18,0.64}

\newcommand{\band}{\wedge}

\newcommand{\fone}{\ensuremath{F_1}}

\newcommand{\xh}{\hat{x}}

\newcommand{\xs}{\bx}

\newcommand{\lvalue}{\textsc{value}}

\newcommand{\lsyn}{\textsc{equivalence}}
\newcommand{\lana}{\textsc{analogy}}
\newcommand{\lfact}{\textsc{fact}}

\begin{document}
\maketitle
\begin{abstract}
\looseness=-1 
To understand a sentence like ``whereas only 10\% of White Americans live at or below the poverty line, 28\% of African Americans do'' 
it is important not only to identify individual facts, e.g., poverty rates of distinct demographic groups, but also the higher-order relations between them, e.g., the disparity between them.
In this paper, we propose the task of Textual Analogy Parsing (TAP) to model this higher-order meaning.
The output of TAP is a frame-style meaning representation which explicitly specifies what is shared (e.g., poverty rates) and what is compared (e.g., White Americans vs.\ African Americans, 10\% vs.\ 28\%) between its component facts.
Such a meaning representation can enable new applications that rely on discourse understanding such as automated chart generation from quantitative text.
We present a new dataset for TAP, baselines, and a model that successfully uses an ILP to enforce the structural constraints of the problem.

\end{abstract}

\section{\label{sec:intro_v3}Introduction}

The task of information extraction by and large seeks to populate a knowledge base with individuated facts extracted from text \citep{sarawagi2008information}. For example, given the sentence:
\begin{quote}
  \leavevmode\llap{\textbf{(E1)} }[According to the U.S. Census, whereas only 10\% of White Americans live at or below the poverty line today]$_{\textrm{C1}}$, [28\% of African Americans do.]$_{\textrm{C2}}$\footnote{Data in E1 and the figure sentence  from \citet{morris2014blackstats}.}
\end{quote}
one would extract two independent facts about voter registration, about the two distinct demographic groups. 
On the other hand, the theory of discourse maintains that part of the above sentence's meaning inheres in the fact that clauses C1 and C2 are juxtaposed \citep{kehler2002}. 
Thus the author intends that we consider them in relation to each other, inviting us to note, for example, a disparity of wealth distribution \textit{between} demographic groups.
To fail to capture this is to miss out on an important aspect of text understanding.


\begin{figure}[t]
  \centering
  \includegraphics[width=\columnwidth]{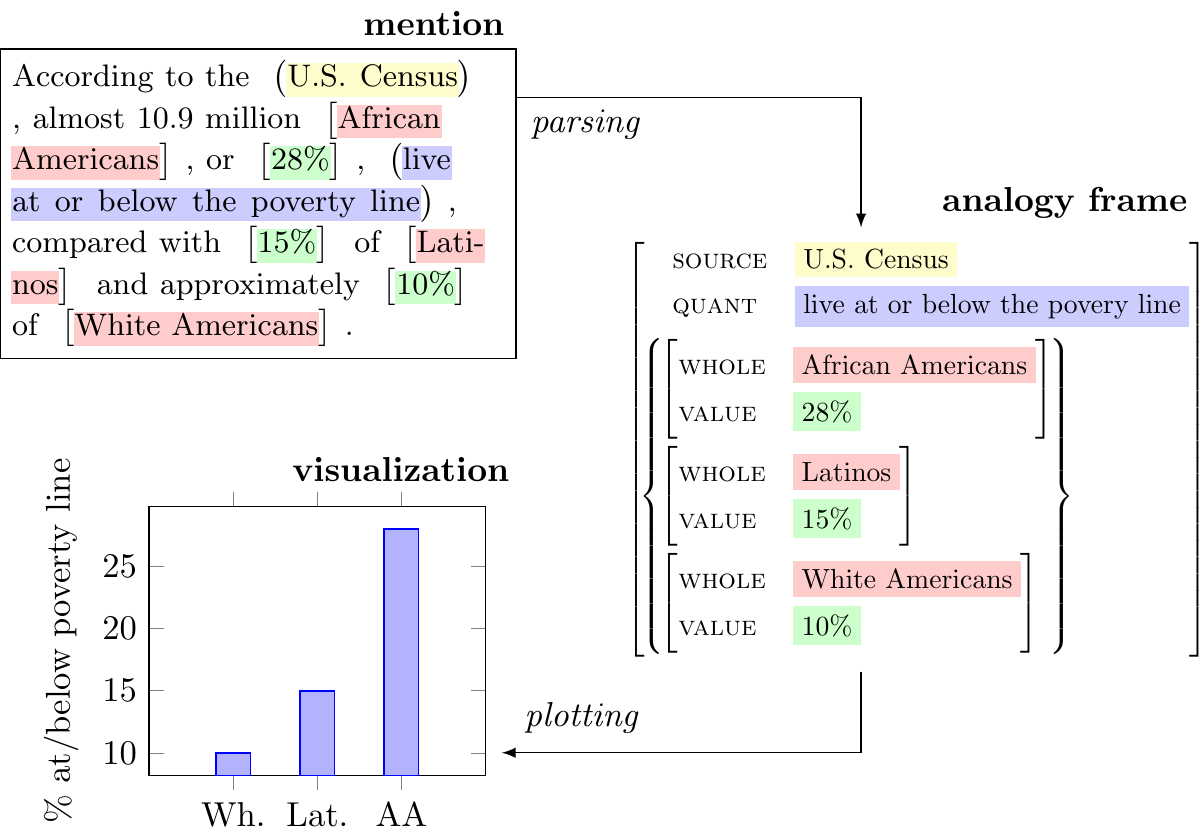}
  \caption{\label{fig:overview} In textual analogy parsing (TAP), one maps analogous facts to semantic role representations and identifies analogical relations between them. Automated chart generation from text is a motivating application of TAP\@.
  }
\end{figure}
\begin{figure*}[ht!]
  \centering
  \includegraphics[width=\textwidth]{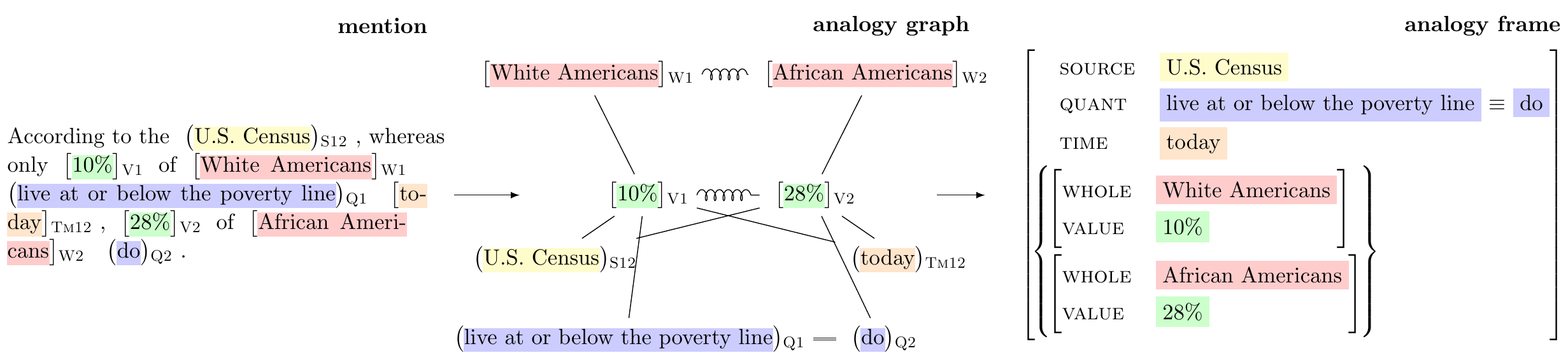}
  \caption{\label{fig:qtap}
 The mapping from utterance to TAP frame.
  Vertices in the graph are labeled with abbreviated semantic roles.
  Single lines represent edges between a \textsc{value} and other roles in its associated fact.
  Double lines represent coreference and synonymy.
  Springs represent analogy.
  Note that vertices connected by equivalence arcs, or any span which connects to both \textsc{V1} and \textsc{V2} via fact relations (i.e., scope), map to the \textit{shared content} of the TAP frame.
  Analogous spans map to the \textit{compared} content.}
\end{figure*}


We propose the task of Textual Analogy Parsing (TAP) to explicitly capture such relational meaning between analogous facts in text.
Concretely, TAP first maps a set of analogous facts to semantic role (SRL) representations, and then identifies the roles along which they are similar (the shared content) and along which they are distinct (the compared content)---see \reffig{overview}. The resulting representation, the TAP frame, is a deeper representation than the one output by shallow discourse parsers~\citep{taboada2006rhetorical,prasad2007penn,pitler2009automatic,prasad2010exploiting,surdeanu2015two}. Given (E1) above, a shallow discourse parser would classify the relation of contrast between C1 and C2---indicating that some salient differences exist in the meanings of the juxtaposed phrases---but without identifying the nature of those differences.

We focus on applying TAP to quantitative facts, because TAP frames can be used to create graphical plots from sentences with numbers, as in \reffig{overview}. This new application could help to simplify complex quantitative text on the web~\cite{barrio2016improving,nytInternal2017}. We thus created an expert-annotated dataset of TAP frames over quantitative facts in the Wall Street Journal corpus~\citep{treebank3}. 

We model TAP by jointly predicting SRL representations of facts in a sentence, and higher-order semantic relations between them.  Our main findings are that a neural architecture outperforms a log-linear baseline, well-chosen linguistic features help performance, and so does the use of an integer-linear programming (ILP) decoder that enforces the structural constraints of the task. Nevertheless, both quantitative and qualitative evaluation reveal  room for improvement on TAP\@.


In sum, our main contributions are (1) a new task, Textual Analogy Parsing (TAP), that combines shallow semantic parsing with discourse meaning, (2) a dataset of TAP frames from quantitative newswire, and (3) a preliminary study of a new application, automated chart generation from text. 
All data and code, including standardized evaluation scripts, are made freely available. 

\section{\label{sec:semantics}A Semantic Representation of Analogy}

\begin{figure}
\centering
\includegraphics[width=\columnwidth]{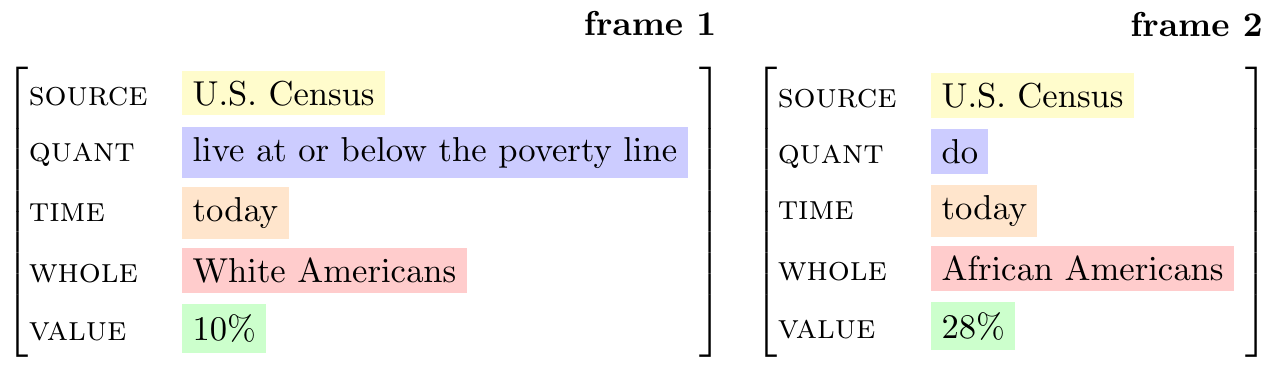}
\caption{\label{fig:facts} Two analogous quantitative facts represented independently, using the QSRL schema~\citep{lamm2018qsrl}.}
\end{figure}

Let us revisit the example sentence from the previous section (E1),
where a pair of analogous quantitative facts about poverty rates of different demographic groups are presented in contrast. Individually, these can be represented using the semantic role structures in~\reffig{facts}, but representing them separately in this way fails to capture the fact that they are analogous, i.e., structurally and semantically similar but distinct.

Instead, we can explicitly show points of similarity and difference between them in the two-tiered frame structure in \reffig{qtap}, which we call a TAP frame. The outer tier of the TAP frame contains \textit{shared content}, or information pertinent to all of the facts in question, and the inner tier contains \textit{compared content}, the information that varies across the set of facts.

Mapping from an utterance to a TAP frame requires three types of relational reasoning. 
Firstly, one must decompose the utterance into a set of facts, where a fact is represented as a set of semantic roles. Then, one must identify the shared content across facts by aligning roles that are semantically equivalent, in the sense that they are either the same span, are coreferent, or are synonymous. For example, in~\reffig{qtap} the phrase `U.S. Census' occurs as the \textsc{source} of both facts because it scopes over the entire sentence in which they appear. Additionally, one must identify the compared content by aligning roles that are analogous, in the sense that they are semantically similar but nevertheless distinct. For example, the  phrases `White Americans' and `African Americans' are analogous in our running sentence, playing the same role in their respective facts, while signifying distinct demographic groups.

\begin{table*}
\centering
\input{examples.table}
\caption{\label{tab:annotatedExamples}
Representative sentences from the Quantitative TAP dataset.
Co-indexing (e.g., \textsc{A1}/\textsc{Q1}) indicates when spans are part of the same QSRL fact. Parentheses indicate shared content spans and brackets indicate compared content spans. To parse (a), one must recognize that `\textit{to acquire PS of New Hampshire}' is elided but nevertheless an implied \textsc{th}(eme) in two of the clauses, and that `\textit{offered}' and `\textit{bid}' are contextually synonymous \textsc{q}(uantities). Moreover, one must note that the \textsc{A}(gents) are analogous, and hence part of the compared content. In (b), `\textit{First Boston}', `\textit{UAL}' and `\textit{worth}', contribute a \textsc{s}(ource), \textsc{th}(eme), and \textsc{q}(uantity) to the shared content respectively. Here, \textsc{c}(ause) roles are compared content.}

\end{table*}

\section{\label{sec:dataset} The Quantitative TAP Dataset}

Motivated by the application of automated graphical plot generation from text, we annotated a dataset of quantitative TAP frames from the Penn Treebank WSJ corpus~\citep{treebank3}.

As our SRL representation of quantitative facts, we employ the Quantitative Semantic Role Labeling (QSRL) framework we previously defined in \citet{lamm2018qsrl}.
Having identified a numerical \textsc{value} in text (e.g., 10\%), QSRL asks, ``what does this number measure?'' to determine its associated \textsc{quantity} (e.g., a poverty rate).
It might also identify, for example, the \textsc{whole} out of which this percentage is measured (e.g., the set of African Americans), and the \textsc{time} at which the quantity took on the value (e.g., today), etc. We employ all fifteen QSRL roles in our annotations.

\begin{table}[t]

\centering
\begin{tabular}{l r r r @{\hspace*{1.2em}} r r r} \toprule
    & \multicolumn{3}{l@{\hspace*{1.2em}}}{Train ($n=1000$)} & \multicolumn{3}{l}{Test ($n=100$)} \\ 
    & av. & max & tot. & av. & max & tot.\\
    \midrule
     Count &
     1.4 & 3 & 1383 &
 1.4 & 3 & 133 \\
   Length &
     2.6 & 16 & -- &
     2.6 &  7 & -- \\
    \bottomrule
\end{tabular}

\caption{\label{tab:data_stats} Dataset statistics (average per sentence, max per sentence, and total over the dataset) for the number of analogy frames (Count) and the number of values compared within each frame (Length).}
\end{table}

Our annotations not only capture the relation between a quantitative predicate and its arguments, but also the higher-order analogy relations between them. The distinction is reflected in the sentences in \reftab{annotatedExamples} from the dataset: Colored spans are co-indexed when they participate in the same quantitative fact; spans with like roles surrounded by parentheses are shared content, meaning that they are either synonymous or co-referent; spans with like roles surrounded by brackets are compared content, meaning that they are analogous but semantically distinct.

To identify instances of quantitative analogy in the WSJ corpus, we first prune out any sentence having fewer than three numerical mentions, where a numerical mention is defined as a contiguous sequence of \texttt{CD} POS tags. Of those left, we manually identify those containing one or more quantitative analogies, i.e., ones in which numerical values are compared content. We estimate the incidence of these to be around 20\%. A linguist then annotated 1,100 of these for analogy relationships. See \reftab{data_stats} for a summary. 

Using an independent set of expert annotations on 100 of these sentences, we measured a significant per-token label agreement of $0.882$ and edge label agreement of $0.991$ using Krippendorf's $\alpha$.\footnote{High edge agreement should be expected because edges are type-constrained and thus easy to identify.
Additionally, we computed agreement after matching overlapping spans.} 

\reftab{annotatedExamples} highlights some of the challenging linguistic phenomena in the data. With respect to identifying the shared content of a TAP frame, these can be coarsely divided into two sets. Firstly, in scope, ellipsis, and gapping, a single syntactic element serves as a role in multiple QSRL frames. This is exemplified by the phrase `PS of New Hampshire' in \reftab{annotatedExamples}(a): It is mentioned explicitly as a \textsc{theme} of the first fact, and only implied in the second two. Based on a random sample of 100 train sentences, we estimate that 86\% of frames in the data exhibit these phenomena. Secondly, in synonymy and coreference, multiple elements appear in a sentence but contribute the same role to the shared content, e.g., `offered' and `bid' in \reftab{annotatedExamples}(a). We estimate that 31\% of frames in the data exhibit these phenomena.

One must learn to identify analogy relationships over a diverse set of compared content roles, with distinct semantic properties: in \reftab{annotatedExamples}(a), \textsc{agent} is a compared content role, whereas in \reftab{annotatedExamples}(b), \textsc{cause} is.

\section{Modeling TAP in the Quantitative Setting}
\label{sec:pipeline}



We model TAP by generating a typed analogy graph over spans of an input text that is isomorphic to the set of TAP frames in that text, e.g., \reffig{qtap}. Each vertex in the graph corresponds to a role-labeled span, and edges represent semantic relations between them.

In this graph, each fact is uniquely identified by a \textsc{value} vertex, which is connected via a \textsc{fact} edge to all of its associated roles. Any two shared content vertices across facts are connected by an \lsyn{} edge, indicating that they are coreferent or synonymous. A single vertex can also be shared across facts by linking via a \textsc{fact} edge to more than one \textsc{value} vertex, suggesting a scopal relationship. Finally, any two vertices which are compared content in the graph are linked via an \textsc{analogy} edge.

More formally, given an utterance $\xs$ with tokens $x_1, \ldots, x_n$, let $G$ be a graph with vertices $V$ and edges $E$. For a vertex $v=(i,j,l)\in V$, $1\leq i<j\leq n$ are the start and end token indices of a span in $\xs$ with role $l\in \mathcal{L_{Q}}\eqdef\{\textsc{value},\dots,\textsc{quant}\}$, 
the set of QSRL roles. For an edge $e=(v,v',l)\in E$, $v,v'\in V$ and $l\in\mathcal{L_{R}}\eqdef \{\textsc{fact},\textsc{equivalence},\textsc{analogy}\}$.

For $G$ so defined to encode a set of valid TAP frames, it must satisfy certain constraints:
\begin{enumerate}
\itemsep 1pt
  \item \textbf{Well-formedness constraints.} For any two vertices $v,v'\in V$, their associated spans must not overlap.   Furthermore, every vertex must participate in at least one \lfact{} edge, i.e., no disconnected vertices.
  \item \textbf{Typing constraints.} \lfact{} relations are always drawn from a \textsc{value} vertex to a non-\textsc{value} vertex. 
  \lana{} and \lsyn{} are only ever drawn between two vertices of the same role.
  \item \textbf{Unique facts.} If a \lvalue{} vertex $v$ is connected to two distinct vertices $v'$ and $v''$ of the same role via a \lfact{} edge, then $\lsyn(v',v'')$ exists.
  \item \textbf{Transitivity constraints.} \lana{} and \lsyn{} edges are transitive: if $\lsyn{}(v,v')\in E$ and $\lsyn(v', v'') \in E$ then $\lsyn(v, v'') \in E$ also. This also holds for \lana{} edges, but only when $v,v'\mbox{ and }v''$ are \lvalue{} vertices.
\end{enumerate}
\begin{enumerate}[resume]
  \item \textbf{Analogy.} There must be at least one pair of analogous \lvalue{} vertices, and for each such pair, there must be a pair of analogous facts connected to them: if $v, v'$ are two $\lvalue{}$ vertices with $\lana(v,v') \in E$, then there must also exist $w, w'$ as two non-\lvalue{} vertices with $\lfact(v, w) \in E$, $\lfact(v', w') \in E$, $\lana(w, w') \in E$.
\end{enumerate} Note that while these constraints rely on the choice of \textsc{value} as the role that grounds quantitative facts, they reflect the general idea that analogy is a structured mapping between meaning representations.

\section{A Neural and ILP Model for TAP}
\label{sec:local}\label{sec:neural}

\begin{figure}\centering
\includegraphics[width=0.75\columnwidth]{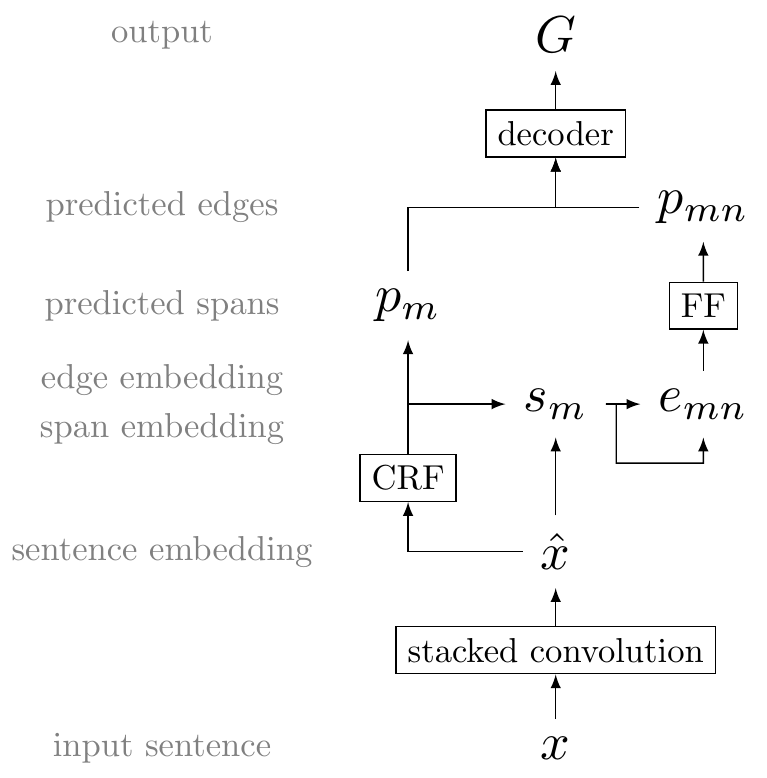}
\caption{\label{fig:nn-overview}
An overview of the proposed neural model:
The sentence embedding represents features across the entire sentence using multiple convolutional layers.
We then use a conditional random field (CRF) layer to predict labeled spans $p_m$ and to generate span and edge embeddings.
We use a feedforward (FF) layer on the edge embeddings to predict edge labels $p_{mn}$.
Together, $p_m$ and $p_{mn}$ form a distribution over edges and labels that we decode into TAP frames.}
\end{figure}

We now present a neural and ILP model that predicts analogy graphs
as defined in \refsec{pipeline}.
Given a sentence, the neural model predicts a distribution over role-labeled spans with edges denoting semantic relations between them. Then, we use an ILP to decode while enforcing the TAP constraints defined in \refsec{pipeline}. \reffig{nn-overview} presents an overview of the architecture.

\paragraph{Context-sensitive word embeddings.}
We first encode the words in a sentence by embedding each token using fixed word embeddings. We also concatenate a few linguistic features to the word embeddings, such as named entity tags and dependency relations. These features are generated using CoreNLP~\citep{manning2014stanford} and represented by randomly-initialized, learned embeddings for symbols together with the fixed word embedding of each token's dependency head and the dependency path length between adjacent tokens. 
The token embeddings are then passed through several stacked convolutional layers~\cite{kim2014convolution}.
While the first convolutional layer can only capture local information, subsequent layers allow for longer-distance reasoning.

\paragraph{Span prediction.}
Next, we feed the outputs of a single fully-connected hidden layer to a conditional random field (CRF)~\citep{lafferty2001conditional}, which defines a joint distribution over per-token role labels. We thus obtain spans from this distribution corresponding to vertices of the graph described in \refsec{pipeline} by merging contiguous role-labels in the maximum likelihood label sequence predicted by the CRF\@.

\paragraph{Edge prediction with \textsc{PathMax} features.}

For edge prediction, we use the spans identified above to construct span and edge embeddings:
  for every span $(i,j)$ that was predicted, we construct a span vector $s_m = \sum_{k=i}^j \xh_{k}$. We also construct a role-label score vector for the span, $p_m$ by summing the role-label probability vectors of its constituent tokens.
  Then, for every vertex pair $(m, n)$, we construct an edge representation $e_{mn}$. The basis of this representation is simply the concatenation of the span representations, the sum of the span representations, their respective role-label score vectors $p_m$ and $p_n$, and relative token distances. 

To capture long-distance phenomena like scope, we also incorporate features into $e_{mn}$ from the dependency paths between the two spans by 
max-pooling the (learned) dependency relation embeddings along the path between the tokens.\footnote{%
  The dependency paths are directed but unlexicalized.
  }
When computing the representation between two spans, we take the average of the path embedding between each pair of tokens within them.
We call this extension \textsc{PathMax}.

The resulting edge representation $e_{mn}$ is passed through a single fully-connected hidden layer and an output layer to predict a distribution over edge labels $p_{mn}$, for each pair of spans.

\paragraph{Training.}

The supervised data described in \refsec{dataset} provides gold spans and edges between them. Thus we define a loss function with two terms: one for the log-likelihood of the span labels output by the CRF model, and one for the cross-entropy loss on the edge labels. We train the span and edge components of the model jointly.

\paragraph{Decoding.} We consider two methods for decoding the span-level and edge-level label distributions $p_{m}$ and  $p_{m,n}$ into a labeled graph respecting the constraints described in~\refsec{pipeline}.

As a simple greedy method to enforce these constraints, we begin by picking the most likely role for each span and edge and then discarding any edges and spans that violate the well-formedness (1) and typing constraints (2).
We then enforce transitivity constraints (4) by incrementally building a cluster of analogous and equivalent spans.
We then resolve the unique facts constraint (3) by keeping only the span with highest \textsc{fact} edge score.
Finally, for every cluster of analogous \textsc{value} spans, we check that the analogy constraint (5) holds and if not, discard the cluster.

We also implement an optimal decoder that encodes the TAP constraints as an ILP \citep{roth2004linear,do2012joint}. The ILP tries to find an optimal decoding according to the model, subject to hard constraints imposed on the solution space. For example, we require that solutions satisfy the `connected spans' constraint: $$\forall s \exists s': e(s,s',\lfact{})$$
\noindent In plain English, this says that every span $s$ in a solution must be connected via a \textsc{fact} edge to some other span $s'$. See the supplementary material for the full list of  constraints we employ. We solve the ILPs with Gurobi~\cite{gurobi}.

\section{Experiments}
\label{sec:evaluation}

We now describe the experimental setup of our neural model (\refsec{local}) on the dataset of TAP frames we created (\refsec{dataset}). Results and discussion are reported in~\refsec{discussion}.

\paragraph{Evaluation metrics.}
The primary metric we use to measure the accuracy of a system on frame prediction is the precision, recall and \fone{} between the labeled vertex-edge-vertex triples predicted by the model and those in the gold parse.
If there are multiple predicted spans that overlap with a single gold span or vice versa, we find a matching of predicted and gold spans that maximizes overlap.

In addition to the primary metric, we also report precision, recall and \fone{} when predicting labeled (non-\textsc{value}) spans and predicting labeled edges before performing any decoding.\footnote{%
We exclude \textsc{value} spans from span scores because they are easy to predict and thus inflate model performance.}
We also use the matching process described above for both these sets of metrics.
Standardized evaluation code is provided with the dataset.

\paragraph{Experimental setup.}
We compare the neural models presented in \refsec{local} in addition to a log-linear baseline.
The log-linear baseline uses the same fixed word embeddings as the neural model in addition to the named entity and dependency parse features described in \refsec{local}.
The key difference is that instead of learning a sentence embedding or hidden layers, the log-linear model simply uses a CRF to predict span labels directly from fixed input features, and then uses a single sigmoid layer to predict edge labels from deterministic edge embeddings, $e_{mn}$.

For the neural models, we used three convolutional layers for sentence embedding with a filter size of 3.
Every layer other than the input layer used a hidden dimension of 50 with ReLU nonlinearities.
We introduced a single dropout layer ($p=0.5$) between every two layers in the network (including at the input).
We used 50-dimensional GloVe embeddings \citep{pennington2014glove} learned from Wikipedia 2014 and Gigaword 5 as pre-trained word embeddings, and initialized the embeddings for the features randomly. 
We chose relatively low input- and hidden-vector dimension because of the size of our data.
The network was trained for 15 epochs using ADADELTA~\citep{zeiler2012adadelta} with a learning rate of $1.0$.
All models were implemented in \texttt{PyTorch}~\citep{paszke2017automatic}.

\begin{table}[t]
    \centering
\begin{tabular}{l l l r r  >{\columncolor[gray]{0.9}}r} \toprule
& & & \multicolumn{3}{c}{Frame prediction}\\ \cmidrule{4-6} 
Model & Feats. & Dec. & P & R & \fone{} \\ 
\midrule
Log-linear & $\checkmark$ & gr. & 
  46.3  &  21.8  &  29.7 \\
Log-linear & $\checkmark$ & opt. & 
  37.1  &  27.5  &  31.6 \\
Neural & $\times$ & gr. & 
  50.7  &  38.4  &  43.7 \\
Neural & $\times$ & opt. & 
  52.8  &  48.6  &  50.6 \\
Neural & $\checkmark$ & gr. & 
  54.9  &  57.4  &  56.1 \\
Neural &  $\checkmark$ & opt. & 
  \textbf{56.4}  &  \textbf{68.8}  &  \textbf{62.0} \\
       \bottomrule
\end{tabular}

    \caption{\label{tab:frame} Performance of models on the test data. Combining the neural model with linguistic features and using an optimal decoder to enforce semantic constraints led to the best performance. 
    }
\end{table}

\section{\label{sec:discussion} Results and Discussion}

Frame prediction results on the test set are summarized in \reftab{frame}. Our three main findings are that (i) the neural network model far outperforms the log-linear model on our frame metric, (ii) including linguistic features further increases performance, and (iii) so does using an optimal decoder over a greedy method.

\begin{table}[t]
    \centering
\begin{tabular}{l r r  >{\columncolor[gray]{0.9}}r}
\toprule
         & \multicolumn{3}{c}{Span prediction} \\ 
         \cmidrule{2-4}
Model   & P & R & \fone{} \\ \midrule
Log-linear (all feats.) & 
  \textbf{42.8}  &  \textbf{82.3}  &  \textbf{56.3} \\
Neural (no feats.) & 
  41.7  &  79.1  &  54.6 \\
Neural (all feats.) & 
  41.5  &  79.2  &  54.4 \\
\quad w/o NER & 
  41.6  &  79.1  &  54.5 \\
\quad w/o dep. & 
  41.2  &  77.5  &  53.8 \\
\quad w/o CRF & 
  36.1  &  73.1  &  48.3 \\
\bottomrule
\end{tabular}

    \caption{\label{tab:span} Performance of models on labeled (non-\textsc{value}) span prediction during cross-validation prior to decoding.
    We found using a CRF to be the most important aspect: simply using fixed word vectors with a CRF (i.e., the log-linear model) was sufficient to predict spans.} 
\end{table}

\begin{table}[t]
    \centering
\begin{tabular}{l r r  >{\columncolor[gray]{0.9}}r}
\toprule
         & \multicolumn{3}{c}{Edge prediction} \\ 
         \cmidrule{2-4}
Model   & P & R & \fone{} \\ \midrule
Log-linear (all feats.) & 
  33.6  &  15.7  &  18.9 \\
Neural (no feats.) & 
  73.7  &  65.8  &  68.7 \\
Neural (all feats.) & 
  74.4  &  \textbf{75.6}  &  \textbf{74.7} \\
\quad w/o NER & 
  74.8  &  72.2  &  73.1 \\
\quad w/o dep. & 
  73.4  & 65.0 & 68.2 \\
\quad w/o \textsc{PathMax}. & 
  72.8  &  64.0  &  67.2 \\
\bottomrule
\end{tabular}


    \caption{\label{tab:edge} Performance of models on labeled edge prediction during cross-validation prior to decoding.
    We found that both dependency label (dep.) and path features (\textsc{PathMax}) help significantly.
   }
\end{table}

\paragraph{Quantitative error analysis.} To better understand which aspects of our model contribute to the task, we perform an ablation study on the span and edge predictions of our model \textit{prior to decoding}.

With respect to span prediction (\reftab{span}), we found that the fixed word vectors, along with a CRF, were able to capture the information needed to identify QSRL role-spans.
Indeed, the log-linear baseline, which directly uses these word vectors as features for a CRF, did the best at span prediction.
We believe that the drop in performance from introducing hidden layers with the neural models is a result of the model updating its span representations to do better edge prediction.\footnote{In a separate experiment, the neural model outperformed the log-linear model when they were trained only to do span prediction.}

While the log-linear model did well at predicting spans, it did a poor job predicting edges, indicating that learning to extract higher-order features from learned span embeddings is necessary for identifying semantic relations between them (\reftab{edge}).
We also found that linguistic features were important:
in particular, we found that syntactic features -- the dependency path features (\textsc{PathMax}) and dependency labels --  played a big role in edge prediction, followed by type information from NER tags. 

\paragraph{Qualitative error analysis.} Our model is tasked with jointly identifying QSRL parses of analogous facts in a sentence, and \lana{} and \lsyn{} relations among them. As described in~\refsec{pipeline}, these pieces interact in mutually constraining ways, and thus it is possible for local errors to have global effects on predicted frames.

\begin{figure}[t!]
\centering
\includegraphics[width=\columnwidth]{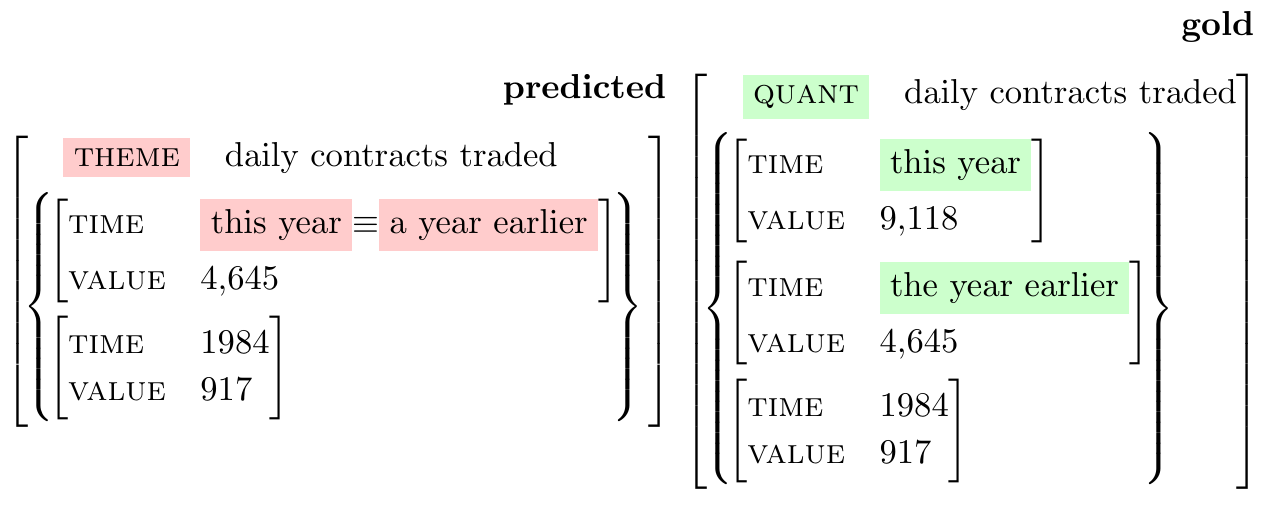}
 \caption{\label{fig:misclassified_equivalence}
TAP frames for the sentence, `\textit{This year \dots daily contracts traded totaled 9,118, up from 4,645 a year earlier and from 917 in 1984.}' The model not only misclassifies the QSRL role of `\textit{daily contracts traded}', but also mistakenly identifies an \lsyn{} between `\textit{this year}' and `\textit{the year earlier}'. As a result, the \textsc{value} {9,118} is left without a compared content role, and is dropped.}
 \end{figure}

In~\reffig{misclassified_equivalence}, for example, the model correctly identifies the gold \textsc{time} spans as part of a TAP frame, but mistakenly predicts that they are linked by \lsyn{}, and thus modify the same \textsc{value} span. In the gold parse, they are linked by \lana{}, and modify distinct \textsc{value} spans. As a result of this misclassification, the model leaves out an entire QSRL fact from the resulting parse.

\begin{figure}[t!]
\centering
\includegraphics[width=\columnwidth]{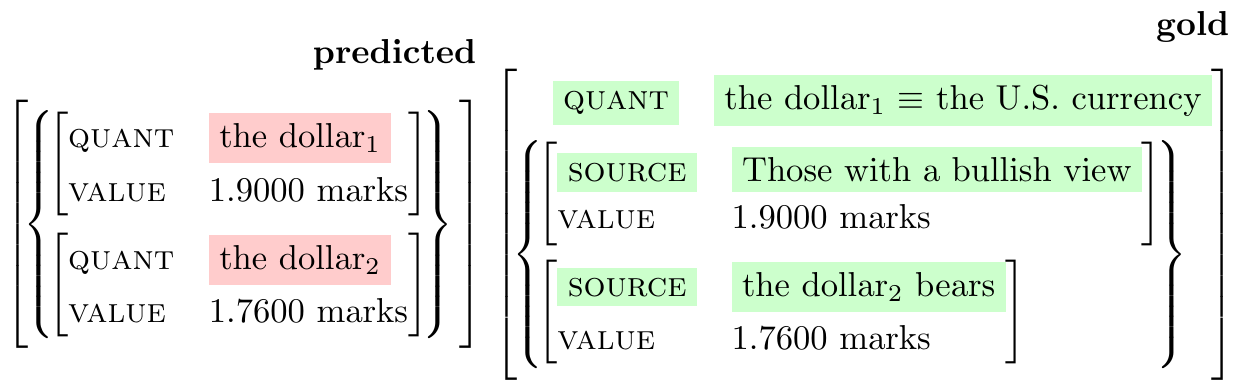}
 \caption{\label{fig:failed_analogy} TAP frames for the sentence `\textit{Those with a bullish view see [the dollar]$_1$ trading up near 1.900 marks\dots while [the dollar]$_2$ bears see the U.S. currency trading around 1.7600 marks}'.
 Among other errors, the model failed to identify analogous \textsc{source} spans and instead predicts that the two instances of the phrase `\textit{the dollar}' (indicated with indexing) in the sentence contribute non-\textsc{value} compared content.}
 \end{figure}

In many cases, the model successfully identifies compared content roles between QSRL facts. In~\reffig{failed_analogy}, we show an example where it does not manage to do so. Here, unable to identify the \lana{} relation between the phrases `\textit{Those with a bullish view}' and `\textit{the dollar bears}', the model instead chooses two identical sequences `the dollar' as the non-\textsc{value} compared content. Inspecting edge probability scores from the model \textit{before decoding} reveals that the neural model thinks that the first instance of `the dollar' in the sentence is semantically analogous to the second; it can be confused by surface similarity into classifying \lana{} relations.
 
\begin{figure*}[!ht]
    \centering
    \begin{subfigure}{0.22\textwidth}
    \includegraphics[width=\textwidth]{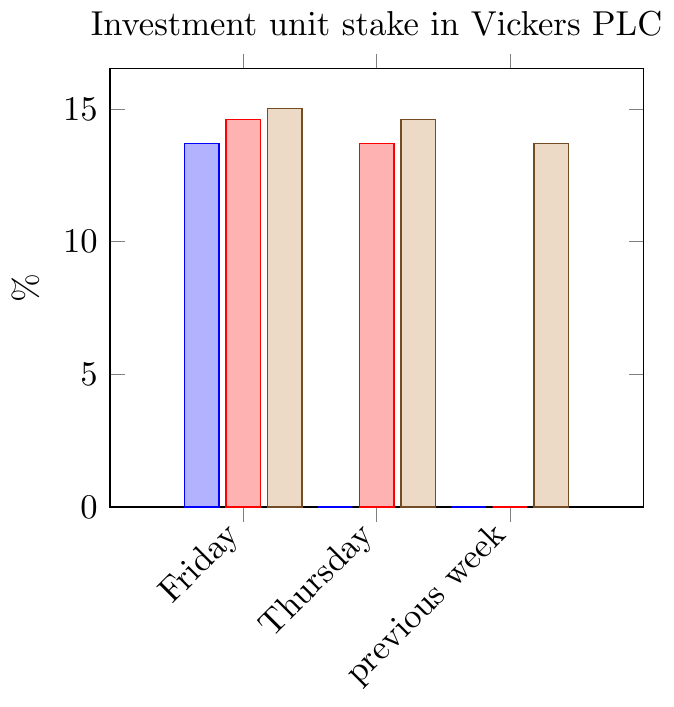}
    \caption{Before constraints}
    \end{subfigure}
    \begin{subfigure}{0.22\textwidth}
    \includegraphics[width=\textwidth]{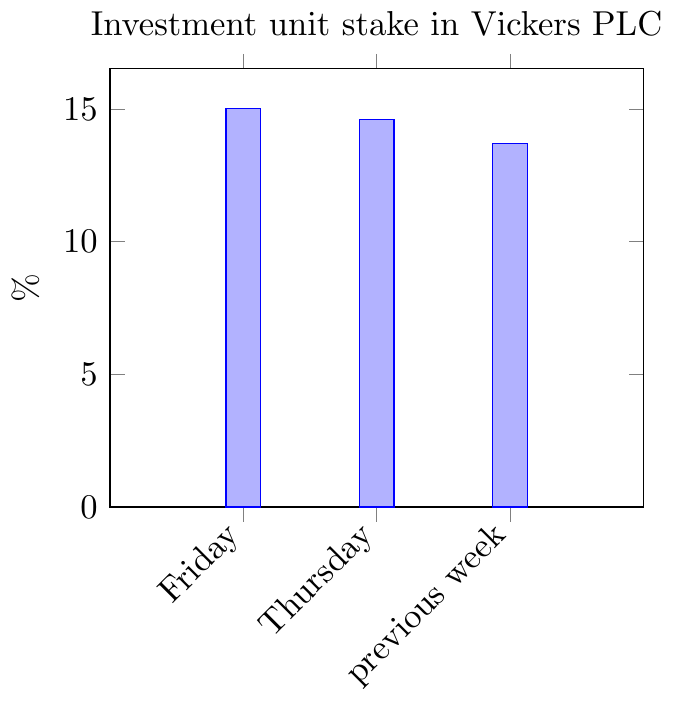}
    \caption{After constraints}
    \end{subfigure}
    \begin{subfigure}{0.22\textwidth}
    \includegraphics[width=\textwidth]{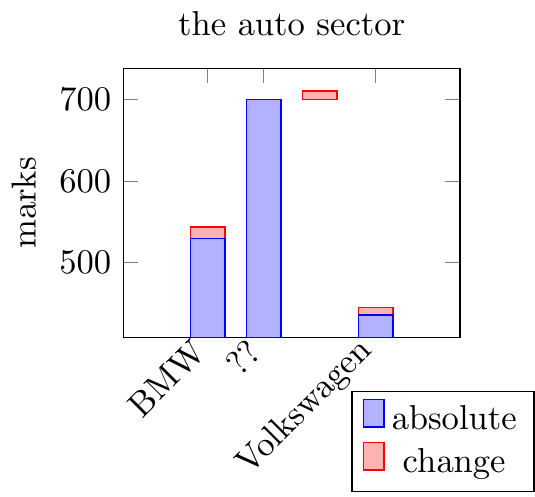}
    \caption{Before constraints}
    \end{subfigure}
    \begin{subfigure}{0.22\textwidth}
    \includegraphics[width=\textwidth]{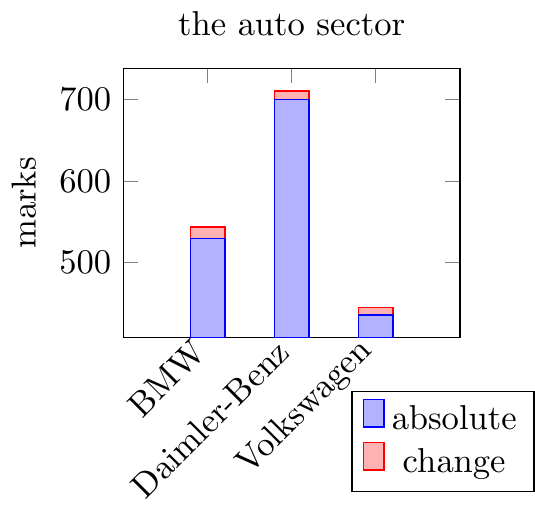}
    \caption{After constraints}
    \end{subfigure}
\caption{\label{fig:plot_results} Charts generated from TAP frames. Charts (a) and (b) are generated from the sentence 	`\textit{Vicker's PLC \dots raised its stake in the company Friday to 15.02\% from about 14.6\% Thursday and from 13.6\% the previous week.}' Before imposing constraints, the neural model assigns multiple values to the \textsc{time} arguments `\textit{Thursday}' and `\textit{Friday}', over-extending their scope. Imposing structural constraints ensures the correct assignment of \textsc{times} to \textsc{values}. Charts (c) and (d) are generated from the sentence `\textit{In the auto sector, Bayerische Motoren Werke plunged 14.5 marks to 529 marks, Daimler-Benz dropped 10.5 to 700, and Volkswagen slumped 9 to 435.5.}' Here, the model fails to associate an absolute (blue) and relative (red) \textsc{value} pair with a \textsc{theme} role. The imposition of global constraints corrects this, linking them to the \textsc{theme} `Diamler-Benz'.}
  \end{figure*}
\paragraph{Application to plot generation.} 
As we have seen, textual analogy is frequently used to compare quantities along some axis of differentiation. For example, one might compare the stock prices of different companies, or describe the change in some quantity's value over time. Such analogy relationships can alternately be expressed in the form of a plot.


Indeed, there is a natural correspondence between charts and TAP frames over quantitative facts: \textsc{values} of a quantitative TAP frame are plotted against other compared content roles, and elements of the shared content correspond with scopal chart elements, such as titles. This mapping is well-defined provided analogous values share units. We present some initial results exploring this direction.

In \reffig{plot_results}, we deterministically plot TAP frames generated by our system both before and after the imposition of global analogy constraints, for two sentences in the data. In the first sentence, \textsc{value} spans are plotted against the \textsc{time} spans the model associates with their respective facts. In the second sentence, two analogy frames are plotted together, one reflecting the absolute values of the stock prices mentioned (blue) and the other reflecting the changes in prices mentioned (red). Units are extracted from \textsc{value} spans using simple pattern matching. Chart titles are only illustrative and were generated by stitching together shared content identified by our system. 

Note that with the imposition of global constraints reflecting the structure of analogy, 
the system yields well-formed charts. Without these constraints, generated charts either have multiple y-axis values assigned to the same x-axis value, or have floating y-axis values with no grounding on the x-axis.

\section{\label{sec:related} Related Work}
\paragraph{Analogy.} In the cognitive science literature, analogy is 
a general form of relational reasoning unique to human cognition~\citep{tversky1978studies,holyoakThagardMentalLeaps,goldstone2005transfer,penn2008darwin,holyoak2012analogy}. Our model of textual analogy is particularly influenced by Structure Mapping Theory \citep{falkenhainer1989structure,gentner1997structure}, an influential cognitive model of analogy as a structure-preserving map between concepts.

Within the NLP community, there has been much work focused on inferring lexical analogies between generic concepts, e.g., \textit{tennis:\linebreak[0]racket::\linebreak[0]baseball:bat} \citep{mikolov2013linguistic,turney2013distributional}, from global distributional statistics. Such analogies are generic, type-level patterns whose structure exists in the nature of the language; here, we are interested in specific analogies whose structure is conveyed by a particular sentence.


\paragraph{Discourse and Information Extraction.} TAP is an information extraction task that synthesizes ideas from semantic role labeling on the one hand and discourse parsing on the other. The former produces predicate-argument representations of individual facts in a text \citep{baker1998berkeley,gildea2002automatic,palmer2005proposition}; the latter identifies discourse relations between syntactic clauses \citep{taboada2006rhetorical,prasad2007penn,pitler2009automatic,prasad2010exploiting,surdeanu2015two}. 

TAP first maps from syntax to a set of SRL-style representations, and then identifies structurally-constrained, higher-order relations among them. It is in this sense reminiscent of, but distinct from, work on causal processes by \citet{berant2014modeling}.

\paragraph{Numbers in NLP.} There has been some work on understanding numbers in text. This includes quantitative reasoning \citep{kushman2014learning,royQuantitiesTACL}, numerical information extraction \citep{madaan2016numerical}, and techniques for making numbers more easily interpretable in text \citep{chaganty2016howmuch,kim2016generating}. 

If pursued further, the application of plotting quantitative text that we discuss in this paper could help to clarify quantitative text on the web \citep{larkin1987diagram,barrio2016improving}.

\paragraph{Neural modeling.} Recent work has shown the promise of sophisticated neural models on semantic role labeling~\citep{he2017deepsrl}.
Similar to other such sequence prediction models, e.g., those for named entity recognition~\citep{lample2016neural} or semantic role labeling~\citep{zhou2015end}, our span prediction utilizes a neural CRF\@.
Our model also has an edge-prediction component, which benefits from a simplified version of the PathLSTM model of \citet{roth2016neural}.
Our edge-prediction model also uses an embedding concatenation component, which was inspired by recent work on neural coreference resolution~\cite{lee2017end}.
\citet{he2017deepsrl} also impose semantic constraints during prediction, but use A$^*$ search instead of an ILP\@.


\section{Conclusion}
\label{sec:conclusion}

In this paper we have presented a new task, textual analogy parsing, or TAP\@. Given a sentence about a set of analogous facts, TAP outputs a frame representation that expresses the points of similarity and difference in their meanings. 

We note that in the particular case of quantitative text, TAP frames correspond with charts. We develop a new dataset of TAP frames from quantitative newswire, and compare a variety models for TAP\@. Our best model employs a globally optimal decoder to enforce the structural constraints of analogy; its outputs can be mapped to well-formed charts of quantitative information extracted from text. 

We view this work to be an exciting step in the direction of deeper discourse modeling. Future work might further extend the recovery of analogy as part of information extraction. This might include TAP outside of the quantitative domain, or TAP at the paragraph level.

\section*{Acknowledgments}

We would like to thank the members of the Stanford NLP Group for reviewing early versions of the paper, and would also like to thank the anonymous reviewers for their thoughtful feedback. 

\bibliography{analogies}
\bibliographystyle{acl_natbib_nourl}

\clearpage
\appendix
\section{\label{appendix:ilp} ILP constraints}

The optimal decoder described in~\refsec{local} implements the TAP constraints in~\refsec{pipeline} as an ILP. Here we present a representative set of our ILP constraints in the form of boolean expressions. In the following, variables $s$ refer to spans in an input text. $\mathcal{L_Q}=\{\textsc{quant}\dots\textsc{value}\}$ is the set of QSRL roles, and $\mathcal{L_{R}}=\{\lsyn{},\lana{},\lfact{}\}$ is the set of semantic relations between them. 

We let $r(s,l) = 1$ if the decoder assigns to $s$ the role-label $l\in\mathcal{L_{Q}}$, and 0 otherwise, and let $\bar{r}(s)=1$ if for any role-label $l\in\mathcal{L_{Q}}$ $r(s,l)=1$, and zero otherwise. Similarly, we let $e(s,s',l)=1$ if an edge is identified between spans $s, s'$  with label $l\in\mathcal{L_{R}}$, and let $\bar{e}(s,s')=1$ if for any edge label $l\in\mathcal{L_{R}}$, and 0 otherwise.

\newcommand{\rb}{\bar{r}}
\begin{enumerate}
\item (Unique Roles) $\forall s: \sum_{l\in\mathcal{L_{Q}}} r(s,l) = \rb(s)$
\item (Unique Edges) $\forall s, s': \sum_{l\in\mathcal{L_{R}}} e(s,s',l) = \bar{e}(s,s')$
\item (Connected Spans) $\forall s \exists s': e(s,s',\lfact{})$
\item (Active Edges) $\forall s, s': \bar{e}(s, s') \implies \rb(s) \band \rb(s')$
\item (Equivalence and Analogy Typing) $\forall l,s, s'$: \begin{itemize}
\item $e(s,s',\lsyn{}) \implies \exists l: r(s,l) \wedge r(s',l)$ 
\item $e(s,s',\lana{}) \implies \exists l: r(s,l) \wedge r(s',l)$
\end{itemize}
\item (Fact Typing) $\forall s,s': e(s,s',\lfact{}) \implies (r(s, \textsc{value}) \band \neg r(s',\textsc{value})) \vee (r(s', \textsc{value}) \band \neg r(s,\textsc{value})) $ 
\item (Equality and Analogy Triangles) $\forall s,s',s'':$
\begin{itemize} 
\item $e(s, s',\lsyn{}) \band e(s', s'',\lsyn{}) \implies e(s, s'',\lsyn{})$ 
\item $e(s, s',\lana{}) \band e(s', s'',\lana{}) \implies e(s, s'',\lana{})$ but only when $r(s,\textsc{value})\wedge r(s',\textsc{value}) \wedge r(s'',\textsc{value})$.
\end{itemize}
\item (Fact-Equality) $\forall l,s,s',s'':$\begin{itemize}
\item $e(s,s',\lfact{}) \band e(s', s'',\lsyn{}) \implies e(s,s'',\lfact{})$
\item $e(s, s',\lfact{}) \band e(s, s'',\lfact{}) \band r(s,l) \wedge r(s',l) \implies e(s', s'')$
\end{itemize}
\item (Analogy Quadrangle) $\forall s,s': r(s, \textsc{value}) \wedge r(s', \textsc{value}) \wedge e(s, s', \lana{})\implies \exists s'', s''': e(s,s'',\lfact{}) \wedge e(s',s''',\lfact{}) \wedge e(s'',s''',\lana{})$
\end{enumerate}

\end{document}